\documentclass[final,1p,times]{elsarticle}
\usepackage{graphicx}

\usepackage{bbm}
\usepackage{color,xcolor}
\usepackage{geometry}
\usepackage{setspace}
\biboptions{sort&compress}
\usepackage{subcaption}
\usepackage{float}
\usepackage{hyperref}
\usepackage{booktabs}
\usepackage{makecell}
\usepackage{multirow}
\usepackage{amssymb}
\usepackage{amsmath}

\journal{Engineering Analysis with Boundary Elements}

\begin{document}

\begin{frontmatter}



\title{Convolutional-neural-operator-based transfer learning for solving PDEs}


\author[a]{Peng Fan} 
\author[a,b]{Guofei Pang\corref{cor1}}
\ead{guofei\_pang@seu.edu.cn}
\cortext[cor1]{Corresponding author}

\affiliation[a]{organization={School of Mathematics},
            addressline={Southeast University}, 
            city={Nanjing},
            postcode={211189}, 
            country={China}}

\affiliation[b]{organization={Key Laboratory of Maritime Intelligent Cyberspace Technology (Hohai University), Ministry of Education},
	addressline={Hohai University}, 
	city={Nanjing},
	postcode={210098}, 
	country={China}}  
\begin{abstract}
Convolutional neural operator is a CNN-based architecture recently proposed to enforce structure-preserving continuous-discrete equivalence and enable the genuine, alias-free learning of solution operators of PDEs. This neural operator was demonstrated to outperform for certain cases some baseline models such as DeepONet, Fourier neural operator, and Galerkin transformer in terms of surrogate accuracy. The convolutional neural operator, however, seems not to be validated for few-shot learning. We extend the model to few-shot learning scenarios by first pre-training a convolutional neural operator using a source dataset and then adjusting the parameters of the trained neural operator using only a small target dataset. We investigate three strategies for adjusting the parameters of a trained neural operator, including fine-tuning, low-rank adaption, and neuron linear transformation, and find that the neuron linear transformation strategy enjoys the highest surrogate accuracy in solving PDEs such as Kuramoto-Sivashinsky equation, Brusselator diffusion-reaction system, and Navier-Stokes equations. Code is available at \!\url{https://github.com/PengFan130/CNO_based_transfer_learning_for_solving_PDEs}.

\end{abstract}


\begin{highlights}
\item Proposing a few-shot convolutional neural operator (CNO) framework. We extend the CNO to few-shot learning scenarios, demonstrating its validity for transferring knowledge of solution operator from a source dataset to a target dataset.
\item Comparing three transfer strategies. We investigate and compare the transfer accuracy of three transfer strategies: fine-tuning, low-rank adaption, and neuron linear transformation (NLT). Our results show that the NLT achieves the highest surrogate accuracy among all strategies.
\item  Performing extensive validation. We benchmark the proposed transfer-learning framework on three challenging physical systems, including the Kuramoto-Sivashinsky equation, the Brusselator diffusion-reaction system, and the Navier-Stokes equations, demonstrating the reliability of the framework in capturing complex spatiotemporal dynamics.
\end{highlights}


\end{frontmatter}


\section{Introduction}
\label{sec:intro}
Deep-learning-based methods have been extensively employed to solution of partial differential equations (PDEs) \cite{karniadakis2021physics, raissi2019physics, pang2019fpinns, lu2021learning, kovachki2023neural, li2020fourier, fu2024physics, zhang2026novel, zhai2023deep, du2025forward}. Among these, neural operators \cite{liu2025architectures} that approximate the solution operators of PDEs have attracted significant attention. Lu et al. introduced the deep operator network (DeepONet), which decomposes the neural operator into a "branch net" for encoding the input function space and a "trunk net" for encoding the domain coordinates \cite{lu2021learning}. Some variants of DeepONet have been recently proposed to address limitations related to data efficiency \cite{peyvan2025fusion}, noise robustness \cite{lin2023b, pensoneault2025uncertainty}, physical consistency \cite{wang2021learning}, and computational scalability \cite{mandl2025separable}. In parallel to spatial domain approaches, frequency domain approaches such as the Fourier neural operator (FNO) \cite{li2020fourier} also gain their popularity. By parameterizing the integral kernel directly in the frequency domain using fast Fourier transforms, FNO achieves global receptive fields and high computational efficiency. Recognizing some of the limitations of the FNO, particularly in capturing transient responses and non-periodic signals, the Laplace neural operator was proposed \cite{cao2024laplace}. Furthermore, to mitigate the aliasing errors caused by not respecting the continuous-discrete equivalence for some neural operators, the convolutional neural operator (CNO) was introduced \cite{raonic2023convolutional}, which provides an alias-free architecture capable of capturing high-frequency details often lost by spectral truncation. In addition to enhancing surrogate accuracy, recent efforts have been made for weakening the dependency on massive labeled datasets. Self-supervised neural operator (SNO) utilizes physics-informed samplers to generate highly accurate training data on-the-fly at a low computational cost \cite{you2025self}, while Jiao et al. introduced a one-shot learning framework leveraging the locality of differential operators to reconstruct global solutions from a single training instance \cite{jiao2025one}. All these advances indicate a paradigm shift from general-purpose approximations to specialized architectures tailored for geometric flexibility \cite{li2309geometry}, physical fidelity \cite{li2024physics}, and algorithmic efficiency \cite{tran2021factorized}.

Although the aforementioned neural operators have achieved certain success in approximating solution operators for PDEs, their accuracy and stability often degrade when applied to unseen physical systems \cite{zhou2024strategies, gupta2022towards}. To address this generalization issue, there has been a growing interest in developing foundation models \cite{ye2024pdeformer, sun2025towards, zhou2025unisolver, bacho2025operator}, inspired by the scaling laws and emerging capabilities of large language models (LLMs). Unlike conventional neural operators tailored to specific PDEs, these foundation models leverage transformer-based architectures to learn unified representations across multiple physical systems. Recent works, such as Poseidon \cite{herde2024poseidon} and GNOT \cite{hao2023gnot}, demonstrate that by pre-training on massive datasets encompassing diverse PDEs, a single model can capture the underlying solution operators shared across different physics. Furthermore, researchers have introduced in-context learning paradigms to operator learning, such as ICON \cite{yang2023context} and PROSE \cite{liu2024prose}, which utilize data prompts or textual descriptions to learn multiple PDEs simultaneously.

Despite conceptually powerful, these foundation models require prohibitive computational resources and massive multi-physics datasets for pre-training. In contrast, practical engineering scenarios often involve a single governing equation in which only specific parameters or geometries vary. As such, \textit{transfer learning} offers a more pragmatic alternative, capable of adapting pre-trained operators to new physical regimes with minimal data and computational cost. A dominant transfer-learning strategy is fine-tuning, where the majority of parameters in a pre-trained model are frozen, and only task-specific layers are updated using data from the target domain \cite{ahn2025deep, xu2025transfer, goswami2022deep}. More recently, inspired by the success of parameter-efficient adaptation in LLMs, low-rank adaptation (LoRA) has emerged as an effective strategy for reducing memory consumption and training cost \cite{zhang2025low}. In particular, Wang et al. applied LoRA to physics-informed neural networks to enable efficient and accurate transfer learning for varied boundary conditions, materials, and geometries \cite{wang2025transfer}.

It is sometimes expensive or time-consuming for practical problems to access high-fidelity simulation data or experimental measurements. This challenge highlights an urgent need for \textit{few-shot transfer learning}, where a neural operator can adapt to new physical problems using only a small number of target samples. The paper presents a systematic framework for CNO-based few-shot transfer learning. To the best of our knowledge, while CNOs have exhibited superior performance in learning alias-free solution operators,  their applicability in data-scarce regimes remains unexplored. We bridge this gap by pre-training a CNO on a source dataset and adapting it to a small target dataset. Specifically, we will evaluate the performance of three transfer strategies: fine-tuning, LoRA, and neuron linear transformation (NLT). The main contributions of this work are summarized below:

\begin{itemize}
    \item  Few-shot CNO framework. We extend the CNO to few-shot learning scenarios, demonstrating its validity for transferring knowledge of solution operator from a source dataset to a target dataset with minimal data dependency.

    \item  Comparison of transfer strategies. We investigate and compare the transfer accuracy of fine-tuning, LoRA, and NLT. Our results show that the NLT achieves the highest surrogate accuracy among all strategies.

    \item  Extensive validation. We benchmark the proposed transfer-learning framework on certain challenging physical systems, including the Kuramoto-Sivashinsky equation, the Brusselator diffusion-reaction system, and the Navier-Stokes equations, demonstrating the reliability of the framework in capturing complex spatiotemporal dynamics.
\end{itemize}

The paper is structured as follows. Section \ref{sec:method} elucidates the problem setting, the CNO architecture, as well as the transfer strategies, namely fine-tuning, LoRA, and NLT to be explored. Section \ref{sec:result} presents the computational examples for three physical systems, in which the performance of the transfer strategies and the sensitivity of the model to the number of few-shot samples are discussed. Section \ref{sec:concluding} concludes the paper.

\section{CNO-based few-shot learning}
\label{sec:method}
\subsection{Problem setting}
\label{subsec31}
Consider a partial differential equation (PDE) governing a physical field $ u(x, t)$ over a spatial domain $x\in \Omega \subset \mathbb{R}^d$ and a temporal interval $t \in [0, T]$ with spatial dimension $d$ and observation time $T$:
\begin{equation}
\begin{array}{ll}
\left\{
\begin{split}
\frac{\partial u(x,t)}{\partial t} & = \mathcal{L}(u(x,t)),\quad (x,t)\in \Omega \times [0,T],\\
u(x,t) & = \phi(x, t), \qquad x\in \partial \Omega, t\in [0,T],\\
u(x,0) & = u_0(x), \qquad x\in \Omega,\\
\end{split}
\right.
\end{array}
\end{equation}
where $\mathcal{L}(\cdot)$ represents a differential operator that involves spatial derivatives of various orders and nonlinear terms, and $\phi(x,t)$ and $u_0(x)$ correspond to boundary and initial conditions, respectively. 

Instead of directly solving the above system for $u(x,T)$ using conventional numerical methods, a neural operator $ \mathcal{S}_{\boldsymbol{\theta}}(\cdot)$ with learnable parameters $\boldsymbol{\theta}$  can be constructed to approximate the mapping from the prescribed initial condition $u_0(\mathbf{x})$ to the solution snapshot $u(\mathbf{x}, T)$ at the observation time $T$:
\begin{equation}
\mathcal{S}_{\boldsymbol{\theta}} : u_0(x) \, \mapsto \, \hat{u}(x,T;\boldsymbol{\theta}) \approx u(x,T)\quad \forall x\in \Omega.
\end{equation}
In the framework of supervised learning, training data should be provided to calibrate the parameters of neural operator, namely $\boldsymbol{\theta}$. Conventional numerical methods are adopted to generate the training set. In other words, given an input function $u_0(x)$, a conventional numerical solver is run to obtain the output function $u(x,T)$ with high solution accuracy. The generating procedure will be run multiple times for $n$ different input functions, so as to attain a training set $D=\{(\mathbf{u}^i_0, \mathbf{u}^i_T)\}_{i=1}^{n}$. Here, the bold-face $u$ denotes a vector or matrix formed by function values evaluated at certain spatial coordinates.

Transfer learning usually occurs when the neural operator is pre-trained on one dataset but adjusted on another dataset. The former dataset is called the \textit{source dataset} (or in-distribution data), while the latter is called the \textit{target dataset} (or the out-of-distribution data). These two datasets could be rather different. The target dataset is usually expensive and hard to obtain in contrast to the source one. We thus expect to learn an approximate PDE-solver by leveraging the information embedded in the source dataset and adjust what we have learned using the very few information hidden in the small target dataset. This is what few-shot learning does. 
In the rest of the paper, we denote the source and target datasets by $D_s = \{( \mathbf{u}_{s_0}^{i}, \mathbf{u}_{s_T}^{i})\}_{i=1}^{n_s}$ and $D_t = \{( \mathbf{u}_{t_0}^{i}, \mathbf{u}_{t_T}^{i})\}_{i=1}^{n_t}$, respectively. Few-shot learning tasks require that the target dataset be much smaller than the source one, i.e., $n_t \ll n_s$.

In this paper, we distinguish $D_s$ and $D_t$ by setting different parameters for generating the input or output functions. For example, we can parameterize the input function as
\begin{equation}
    u_0(x_1,x_2) = \frac{\pi}{K^2} \sum_{i,j=1}^{K} \frac{a_{ij}} {i^2 + j^2} \sin i\pi x_1 \sin j\pi x_2, 
\end{equation}
with random coefficients $a_{ij}$ and a parameter $K$ controlling the complexity of the input function. Taking $K=6$ and $K=12$ yield the input functions for source and target datasets, respectively. The input functions in the target dataset have higher-frequency modes due to a larger $K$. For another example, we can solve Navier-Stokes equations having high and low viscocities to generate output functions $u(x,T)$ for source and target datasets, respectively.

\subsection{Convolutional neural operator (CNO)}
\label{subsec32}
Restricting to a band-limited function space \cite{vetterli2014foundations}, a subspace of Sobolev spaces that consists of functions whose Fourier spectra are compactly supported within a prescribed frequency band:
\begin{eqnarray*}
\mathcal{B}_w(D) = \{ f \in L^2(D) : \text{supp}\hat{f} \subseteq [-w, w]^2 \},
\end{eqnarray*}
where $w > 0$, $D$ is a two-dimensional torus, and $\hat{f}$ represents the Fourier transform of $f$, it can be proven that for any continuous operator $G^\dagger \colon X \to Y$, with Sobolev spaces $X$ and $Y$, there exists an operator $G^* \colon \mathcal{B}_w(D) \to \mathcal{B}_w(D)$ such that $\|G^\dagger - G^*\| < \varepsilon$, for any prescribed accuracy $\varepsilon > 0$ \cite{raonic2023convolutional}. The CNO is a discrete version of $G^*(\cdot)$ , which respects the continuous-discrete equivalence and thus preserves the structures of being resolution-invariant and alias-free \cite{bartolucci2023representation}. 
Specifically, CNO defines a compositional mapping between bandlimited functions \cite{raonic2023convolutional}:
\begin{eqnarray*}
G_{\boldsymbol{\theta}}: u \mapsto P(u) \mapsto v_0(\cdot) \mapsto v_1(\cdot) \mapsto \cdots \mapsto v_L(\cdot) \mapsto Q(v_L).
\end{eqnarray*}
Each layer $v_i(\cdot)$ is the composition of three key manipulations: a convolution, an activation function, and an up- or down-sampling, all of which are designed to preserve band limits. $P(\cdot)$ is a lifting layer to lift input function $u$ to the latent space and $Q(\cdot)$ is a projection operator to project the output in $v_L(\cdot)$ to the output space. The architecture is essentially a modified U-Net, incorporating ResNet and invariant blocks as skip connections to maintain multiscale information flow. The CNN-based architecture requires a fashion of image processing and thus that all values of input and output functions should be formatted in matrices rather than vectors.

With such an architecture, the CNO possesses the universal approximation capability. For a broad class of PDE-defined operators $G^\dagger(\cdot)$, and for any $\varepsilon > 0$, there exists a CNO $G_{\boldsymbol{\theta}}(\cdot)$ satisfying (see Theorem 3.1 of \cite{raonic2023convolutional})
\begin{eqnarray*}
\|G^\dagger(f) - G_{\boldsymbol{\theta}}(f)\|_{L^p(D)} < \varepsilon, \quad p \in \{2, \infty\},
\end{eqnarray*}
for all functions $f$ belonging to bounded subsets of the Sobolev space $H^r(D)$ with sufficient regularity, where $D$ represents a 2D torus.

\subsection{Convolutional neural operator with transfer learning}
\label{subsec33}
In this section, we detail three transfer strategies implemented within the CNO framework. 
\subsubsection{Fine-tuning}
\label{subsec331}
Fine-tuning is a frequently used strategy for model adaptation. The key idea is to freeze the majority of parameters in a well-trained source model and optimize only the remaining parameters on the target dataset, therefore adapting the model to new data distributions while retaining previously learned knowledge \cite{he2021towards}. This strategy has been widely adopted in neural operator frameworks due to its simplicity and effectiveness. In previous studies \cite{chakraborty2021transfer, zhang2025low}, the fine-tuning was typically performed to the last few layers of networks, and we thus in this paper fine-tune only the parameters of last few layers of a CNO decoder.

\subsubsection{LoRA: low-rank adaptation}
\label{subsec332}
Rapid development of LLMs makes the fine-tuning strategy a critical technique for adapting pretrained models to downstream tasks. However, a full-parameter fine-tuning becomes computationally expensive and storage-intensive as the model size increases. To address this challenge, LoRA \cite{hu2022lora} introduces a low-rank decomposition into the parameter updates of a pretrained model. Specifically, given a pretrained parameter-matrix $W_s \in \mathbb{R}^{d \times k}$, the target parameter-matrix $W_t$ is updated by
\begin{equation}
W_t = W_s + \Delta W = W_s + BA,
\end{equation}
where $A \in \mathbb{R}^{r \times k}$ and  $B \in \mathbb{R}^{d \times r}$ are trainable parameters with $r \ll \min(d, k)$. By freezing the pretrained parameters $W_s$ and only optimizing $A$ and $B$ using a target dataset, LoRA effectively reduces the number of trainable parameters from $O(dk)$ to $O(r(d+k))$.

In this study, we extend LoRA to convolutional layers. Specifically, a convolutional kernel $W_t$ of shape $(C, h, w)$ in the target model with channel number $C$, height $h$, and width $w$ is computed as the summation of the convolutional kernel $W_s$ of the same shape in the source model and the product of a low-rank matrix $B$ and a column vector $A$
\begin{eqnarray*}
    W_t &=& W_s + BA\\
    &=& \begin{bmatrix}
\begin{bmatrix} w_{11}^1 & \cdots & w_{1w}^1 \\ \vdots & \ddots & \vdots \\ w_{1h}^1 & \cdots & w_{hw}^1 \end{bmatrix}, \\ \cdots, \\
\begin{bmatrix} w_{11}^c & \cdots & w_{1w}^c \\ \vdots & \ddots & \vdots \\ w_{1h}^c & \cdots & w_{hw}^c \end{bmatrix}
\end{bmatrix}
+
\begin{bmatrix}
    b_{1}^1 & \cdots & b_{r}^1 \\ \vdots & \ddots & \vdots \\
    b_{1}^c & \cdots & b_{r}^c
\end{bmatrix}
\begin{bmatrix}
    a_{1}^1 \\ \vdots \\
    a_{r}^1 
\end{bmatrix} 
\triangleq \begin{bmatrix}
 w_s^1 + \Delta w^1\\ \vdots\\
w_s^c + \Delta w^c
\end{bmatrix}
,
\end{eqnarray*}
where $B \in \mathbb{R}^{C \times r}$ and $A \in \mathbb{R}^{r \times 1}$. Note that the $i$-th component of the resulting vector of $BA\in \mathbb{R}^{C}$ is simultaneously added to all elements of the matrix $w_s^i\in \mathbb{R}^{h\times w}$ for the $i$-th channel with $i=1,2,\cdots,C$. 

\subsubsection{NLT: Neuron linear transformation}
\label{subsec333}
Wang et al. proposed NLT transfer strategy for the cross-domain crowd counting revealing the domain gap \cite{wang2021neuron}. Akin to the LoRA in the preceding section, this strategy modifies the parameters of the source model in a low-cost manner. Instead of adding a low-rank update, it applies a direct affine transformation
\begin{eqnarray*}
    W_t &=& f \times W_s + b \\
    &=& \begin{bmatrix}
f^1 \times \begin{bmatrix} w_{11}^1 & \cdots & w_{1w}^1 \\ \vdots & \ddots & \vdots \\ w_{1h}^1 & \cdots & w_{hw}^1 \end{bmatrix} + b^1\\ \vdots \\
f^c \times \begin{bmatrix} w_{11}^c & \cdots & w_{1w}^c \\ \vdots & \ddots & \vdots \\ w_{1h}^c & \cdots & w_{hw}^c \end{bmatrix} + b^c
\end{bmatrix},
\end{eqnarray*}
with domain factor $f\in \mathbb{R}^{c \times 1}$ and domain bias $b \in \mathbb{R}^{c \times 1}$. Domain factor and bias need to be updated in the transfer stage. This strategy not only inherits the strong expressivity of the source model, but also enables effective generalization to unseen target domains.

\section{Computational examples}
\label{sec:result}
The section implements the transfer learning framework and aims at evaluating the performance of three transfer strategies. All computational examples employ the CNO as the foundational backbone architecture \cite{raonic2023convolutional}. The workflow adopts a standard procedure of \textit{pre-training and transfer}: CNO is first fully trained on a source dataset. Subsequently, the pre-trained model parameters are fully or partially frozen, and the model is adapted to the target dataset using distinct transfer strategies. For benchmarking, we consider three nonlinear PDE systems: the Kuramoto-Sivashinsky equation, the Brusselator diffusion-reaction system, and the Navier-Stokes equations. 

\subsection{Experimental setup and implementation details}
\label{subsec:setup}

Before proceeding on the computational examples, we first introduce the standardized dataset configuration and network hyperparameters used for all examples.

\textbf{Dataset configuration.} 
To emulate a challenging few-shot learning scenario, the size of target dataset is kept small. For the \textit{source dataset}, we utilize a dataset partitioned into 512 samples for training, 128 for validation, and 128 for testing. For the \textit{target dataset}, the training set is limited to only \textbf{16 samples}. However, to ensure a statistically robust performance evaluation, the validation and testing sets for the target dateset are set to have 128 samples each.

\textbf{Architecture and hyper-parameters.} 
For the pre-training stage, we strictly adhere to the network architecture and hyper-parameter settings reported in the original CNO paper \cite{raonic2023convolutional} to ensure baseline consistency. For the transferring stage, optimization is performed using the AdamW optimizer \cite{loshchilov2017decoupled}. The initial learning rate is set to $10^{-3}$ and is dynamically adjusted using a step-decay scheduler (specifically, decaying by a factor of $\gamma = 0.98$ every 10 epochs) to promote stable convergence. The adaptation process is conducted for 500 epochs with a fixed batch size of 16. To alleviate the effect of randomness of initial guesses for the optimizer, we run the same code 5 times and take the average of the computed errors for each run.

\subsection{Kuramoto-Sivashinsky equation}
\label{subsec41}
The Kuramoto–Sivashinsky (KS) equation models the emergence of instability, nonlinear interactions, and chaotic spatio-temporal patterns in physical systems \cite{papageorgiou1991route}. In two spatial dimensions on a periodic unit square $D=[0, 1]^2$, it can be described by the following form, 
\begin{eqnarray}
    \frac{\partial u}{\partial t} = -\frac{1}{2} |\nabla u|^2 - \nabla^2 u - \nabla^4 u.
\end{eqnarray}
The underlying operator $G^\dagger: u_0 \mapsto u(.,T)$, which maps initial condition $u_0$ to the solution $u$ at $T=1$. Initial conditions can be given
\begin{equation}
    u_0(x,y) = \frac{\pi}{K^2} \sum_{i,j=1}^{K} \frac{a_{ij}}{ i^2 + j^2} \sin i\pi x \sin j\pi y, \quad \forall (x,y) \in D,
    \label{eq:init condition}
\end{equation}
with the random coefficient $a_{ij}$ uniformly drawn from $[-1, 1]$. Then high-fidelity reference solutions are produced by using \textit{py-pde} \cite{zwicker2020py} in which finite difference solutions are obtained on a $128 \times 128$ uniform grid. The source dateset and the target dataset are generated by setting $K=6$ and $K=8$ (or $K=12$), respectively. 

\subsection{Brusselator diffusion-reaction system}
\label{subsec42}
The Brusselator diffusion–reaction system describes the formation of dissipative structures in nonlinear chemical kinetics \cite{goswami2022deep}. On the periodic unit square $D=[0, 1]^2$, the governing equations are
\begin{equation}
\left\{
\begin{split}
\frac{\partial u}{\partial t} &= D_0 \nabla^2 u + a - (1 + b)u + vu^2, \\
\frac{\partial v}{\partial t} &= D_1 \nabla^2 v + bu - vu^2,
\end{split}
\right.
\end{equation}
where $D_0=1$ and $D_1=0.1$ denote diffusion coefficients, and $a=1$ and $b=3$ are parameters controlling the reaction kinetics. Our objective is to approximate the solution operator $G^\dagger: v_0 \mapsto v(.,T)$ that maps an initial state $v_0$ to the solution at $T=10$. The following initial conditions are taken:
\begin{equation}
\left\{
\begin{split}
u_0(x, y) & = 1, \\
v_0(x, y) & = \frac{\pi}{K^2} \sum_{i,j=1}^{K} \frac{a_{ij}} {i^2 + j^2}\sin i\pi x \sin j\pi y,
\end{split}
\right.
\end{equation}
with $a_{ij} \sim U[-1, 1]$. The \textit{py-pde} is utilized to obtain numerical solutions on a $128 \times 128$ uniform mesh. As in the section \ref{subsec41}, we generate a source dataset using $K=6$, and a target dataset using $K=8$ or $K=12$.

\subsection{Navier-Stokes equations}
\label{subsec43}
Consider a 2D incompressible Navier-Stokes equations in the streamfunction-vorticity formulation on the periodic unit square $D=[0, 1]^2$. The evolution of the scalar vorticity field $\omega$ is governed by
\begin{equation}\label{NS-eq}
\frac{\partial \omega}{\partial t} +  \mathbf{u}\cdot \nabla \omega =  \nu \nabla^2 \omega.
\end{equation}
Here, $\mathbf{u}$ is the velocity, which is related to the streamfunction $\psi$ via $\mathbf{u}=\left(\frac{\partial \psi}{\partial y},-\frac{\partial \psi}{\partial x}\right)^{T}$. The streamfunction $\psi$ is further related to the vorticity via $-\nabla^2\psi=\omega$.  The parameter $\nu$ in Eq.(\ref{NS-eq}) is the kinematic viscosity. We seek to approximate the operator $G^\dagger: \omega_0 \mapsto \omega(\cdot, T)$, mapping the initial vorticity to the state at $T=3$. The input function $\omega_0$ is sampled from a truncated Fourier series with random Fourier coefficients\cite{koehler2024apebench}. Reference trajectories are generated via the \textit{APEBench} \cite{koehler2024apebench}, which employs a scalable JAX-based spectral solver coupled with exponential time differencing Runge-Kutta integration to ensure high solution accuracy. To assess the model's generalization for different flow regimes, we introduce a distribution shift via the viscosity parameter. Specifically, the source dataset is characterized by a viscosity of $\nu = 5 \times 10^{-4}$, and the target dataset by $\nu = 3 \times 10^{-4}$ (or $\nu = 1 \times 10^{-4}$), thereby introducing more turbulent dynamics and finer-scale structures absent in the pre-training stage.

\subsection{Results}
\label{subsec44}
\subsubsection{Generalization gap for CNO}
\label{subsec441}
Table \ref{Table1} shows the generalization gap for the standard CNO framework. When pre-trained on the source datasets, the CNOs achieve low test errors for all three PDEs. However, when the pre-trained models are directly employed to make predictions for the target datasets without any parameter adaption, the test errors dramatically increase by nearly one order of magnitude. This indicates that the CNO without transfer strategies often fails to generalize well when the distribution of dataset is shifted.

\begin{table}[H]
\centering
\caption{Test errors of the CNOs on source and target datasets in the absence of transfer strategies. The relative $L_1$ test errors in the column of source dataset correspond to the situations where the CNOs are trained and tested on source datasets, while the test errors in the column of target dataset correspond to the situations where the CNOs are trained on source datasets and tested on target datasets. For the source dataset column, the parameter $K$ that controls the complexity of the input function is set to $K=6$ for the Kuramoto-Sivashinsky equation and Brusselator Diffusion-Reaction system, and the PDE parameter $\nu$ that affects the generating of the output function $\omega(x,T)$ is set to $\nu=5\times 10^{-4}$. For the columns regarding the target dataset, the parameter setting is that $K=8$ for Scenario 1 and $K=12$ for Scenario 2 for the Kuramoto-Sivashinsky equation and Brusselator Diffusion-Reaction system, and $\nu=3\times 10^{-4}$ for Scenario 1 and $\nu = 1\times 10^{-4}$ for Scenario 2 for the Navier-Stokes equations. Although the CNOs achieve low test errors and thus generalize well from the training samples to test samples of the source datasets, yet without any parameter adaption on the target datasets they yield larger errors of generalization from the training samples of the source datasets to the test samples of the target datasets, which can be roughly one order of magnitude higher than the test errors for the source datasets.}
\label{Table1}
\renewcommand\arraystretch{1.3}
\small
\begin{tabular*}{\textwidth}{@{\extracolsep{\fill}}lccc}
\toprule
\multirow{2}{*}{\textbf{Equation}} & \multirow{2}{*}{\textbf{Source dataset}} & \multicolumn{2}{c}{\textbf{Target dataset}} \\
\cmidrule(lr){3-4}
 & & \textbf{Scenario 1} & \textbf{Scenario 2} \\
\midrule
Kuramoto-Sivashinsky Equation & 0.1782\% & 1.4520\% & 1.8095\% \\
\midrule
Brusselator Diffusion-Reaction System & 0.4800\% & 1.1254\% & 2.0390\% \\
\midrule
Navier-Stokes Equations & 1.6645\% & 8.6294\% & 11.9493\% \\
\bottomrule
\end{tabular*}
\end{table}

\subsubsection{CNO-based transfer learning}
\label{subsec442}
To reduce the generalization error of CNOs for distribution shift in datasets, we evaluate three transfer strategies, namely fine-tuning, LoRA, and NLT against the standard CNO baseline discussed in section \ref{subsec441}. We can see from Table \ref{Table2} that the transfer strategies can considerably reduce the generalization errors on the target datasets, compared to no transfer cases. In addition, NLT consistently achieves the lowest relative error among all transfer scenarios. Notably, NLT exhibits remarkable resilience to the magnitude of the discrepancy between source and target datasets, which can be quantified by the maximum mean discrepancy (MMD) \cite{gretton2006kernel, gretton2012kernel}. In stark contrast to LoRA and fine-tuning, NLT yields test errors that are insensitive to MMD. Let us take the Kuramoto-Sivashinsky equation as an example. As the transfer learning task shifts from the low discrepancy case ($K=6 \to 8$, MMD = 1.1874) to the high discrepancy case ($K=6 \to 12$, MMD=1.3326) , the LoRA's test error doubles (from 0.80\% to 1.65\%), whereas the NLT's test error remains nearly stable (from 0.39\% to 0.42\%).  

The "Supervised" baseline in the table, trained only on the target dataset without pretraining, yields high test errors (e.g., roughly $14\%$ error for the Navier-Stokes Equations). This failure is attributed to the data scarcity inherent in the few-shot learning setup, which necessitates the transfer learning strategy.  

The snapshots for the approximate and reference solutions for the three PDE benchmarks are presented in \ref{app1}:  Figure \ref{fig2} for the Kuramoto-Sivashinsky equation, Figure \ref{fig3} for the Brusselator diffusion-reaction System, and Figure \ref{fig1} for the Navier-Stokes equations.

\begin{table}[H]
\centering
\caption{Comparison of different transfer strategies for three PDE benchmarks. The distribution shifts are induced by varying the initial condition parameter $K$ and the viscosity $\nu$. Maximum mean discrepancy (MMD) quantifies the magnitude of the distribution shift between source and target datasets. The best and second-best results among transfer strategies are highlighted in \textbf{bold} and \underline{underlined}, respectively. Transfer learning dramatically reduces the test errors on target datasets compared to the case without transfer (i.e. no transfer) as well as the case without pretraining (i.e. supervised). Also, NLT is observed to be the best transfer strategy for these PDEs. }
\label{Table2}
\renewcommand\arraystretch{1.2}
\setlength{\tabcolsep}{0pt}
\footnotesize

\begin{tabular*}{\textwidth}{@{\extracolsep{\fill}}lcccccc}
\toprule
\multirow{4}{*}{} & \multicolumn{2}{c}{\textbf{Kuramoto-Sivashinsky equation}} & \multicolumn{2}{c}{\textbf{Brusselator diffusion-reaction system}} & \multicolumn{2}{c}{\textbf{Navier-Stokes equations}} \\
\cmidrule(lr){2-3} \cmidrule(lr){4-5} \cmidrule(lr){6-7}
 & \multicolumn{2}{c}{$K$} & \multicolumn{2}{c}{$K$} & \multicolumn{2}{c}{$\nu \, (\times 10^{-4})$} \\
 & $6 \to 8$ & $6 \to 12$ & $6 \to 8$ & $6 \to 12$ & $5 \to 3$ & $5 \to 1$ \\
\midrule
\textit{MDD} & 1.1874 & 1.3326 & 0.3567 & 0.3734 & 0.0933 & 0.1617 \\
\midrule
NLT         & \textbf{0.3963\%} & \textbf{0.4232\%} & \textbf{0.6025\%} & \textbf{0.7692\%} & \textbf{3.7113\%} & \textbf{4.9985\%} \\
LoRA        & 0.8079\%          & 1.6562\%          & \underline{0.8750\%}          & \underline{1.1442\%}          & \underline{3.9628\%}          & \underline{5.2407\%} \\
Fine-tuning & \underline{0.7007\%}          & \underline{0.9220\%}          & 0.9166\%          & 1.4474\%          & 4.4574\%          & 6.5024\% \\
\midrule
Supervised  & 0.8599\%          & 1.0991\%          & 1.8588\%          & 2.2783\%          & 13.9494\%         & 16.2400\% \\
No transfer & 1.4520\%          & 1.8095\%          & 1.1254\%          & 2.0390\%          & 8.6294\%          & 11.9493\% \\
\bottomrule
\end{tabular*}
\end{table}

\subsubsection{Sampling efficiency and few-shot learning}
\begin{figure}[H]
\centering
\includegraphics[scale=0.8]{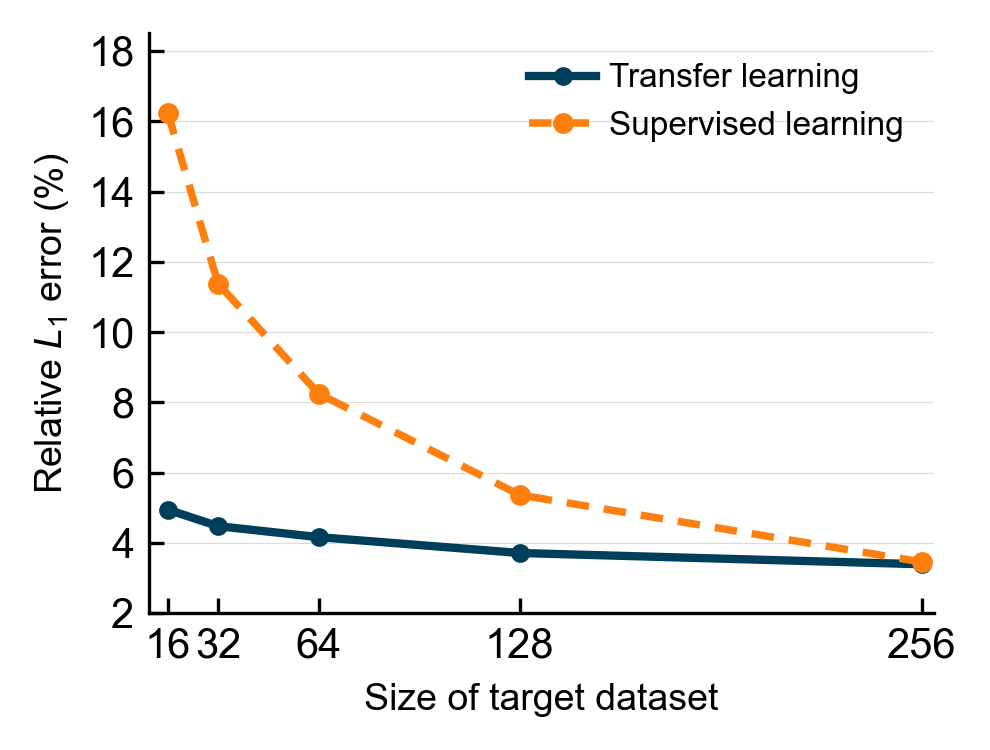}
\caption{Relative $L_1$ error (\%) plotted against the number of samples in the target dataset for the Navier-Stokes equations in which the transfer is made from a high kinematic viscosity to a low one (from $\nu=5\times 10^{-4}$ to $\nu=1\times 10^{-1}$). The blue and orange curves correspond to the transfer learning framework and the supervised learning baseline, respectively. When a large target dataset is available (e.g. for a target dataset of size 256), training a CNO on the target dataset will be enough; however, when the size becomes small (say $n_t=16$) due to the high cost of attaining high-fidelity solutions, the direct training fails, but the transfer learning provides the opportunities for yielding a highly accurate surrogate model.  }\label{fig4}
\end{figure}

We plot in Figure \ref{fig4} the variation of test errors on the size of target dataset $n_t$. We take the Navier-Stokes equations as an example. The source and target datasets are generated for $\nu = 5 \times 10^{-4}$ and $ \nu = 1 \times 10^{-4}$, respectively. The size of target dataset, i.e., $n_t$ varies from 16 to 256. We compare the test error curves for two distinct cases: (1) transfer learning with NLT strategy, and (2) supervised learning directly on the target dataset without pretraining on the source dataset. We can see that the supervised learning baseline has a low test error when the target dataset is sufficiently large ($n_t=256$), but the error dramatically increases as the training dataset size decreases. In particular, at the few-shot regime ($n_t=16$), the error of the supervised model increases to 16.24\%, indicating a complete failure of generalization from the training samples to test samples of the target dataset due to data scarcity. In contrast, transfer learning yields a low test error which is almost insensitive to $n_t$. Although the size is reduced by a factor of 16, the test error increases marginally from $3.4\%$ to $5.0\%$. This implies a critical advantage of transfer learning to leverage the source domain knowledge to compensate for the insufficiency of the target dataset. 

It is also noted that few-shot transfer learning enables a multi-fidelity data fusion, in which high-fidelity PDE dataset is usually small due to the high computational overhead for data generation but low-fidelity PDE dataset can be large because of fast solution procedures.  

\section{Conclusion}
\label{sec:concluding}
In this paper, we propose a convolutional-nerual-operator-based (CNO-based) transfer learning framework as a highly accurate surrogate model for approximating the solution operators of PDEs, particularly when a small number of high-fidelity PDE solutions are accessible. We demonstrate that the standard CNO cannot generalize well from the source dataset to the target dataset when the distributions of the two datasets differ significantly. Incorporating the CNO with transfer strategies such as fine-tuning, low-rank adaption, and neuron linear transformation (NLT) can considerably improve the generalization ability of the CNO. Applying the proposed transfer learning framework to approximation of solution operators for three typical nonlinear PDEs (the Kuramoto-Sivashinsky equation, the Brusselator diffusion-reaction system, and the Navier-Stokes equations), we also observe that the NLT transfer strategy performs best and achieves the highest surrogate accuracy. Our future works include loosening the requirement of a rectangular computational domain and allowing surrogate predictions for time marching.

\section*{Acknowledgments}
This work was financially supported by Jiangsu Provincial Scientific Research Center of Applied Mathematics (BK20233002).

\appendix
\setcounter{figure}{0}
\section{Visualization of computed solutions for PDE benchmarks}
\label{app1}

In what follows, we demonstrate the contour plots for three test samples in the target datasets for the Kuramoto-Sivashinsky equation, the Brusselator diffusion-reaction system, and the Navier-Stokes equations.

\begin{figure}[H]
\centering

\begin{minipage}{\linewidth}
  \centering
  \includegraphics[width=\textwidth]{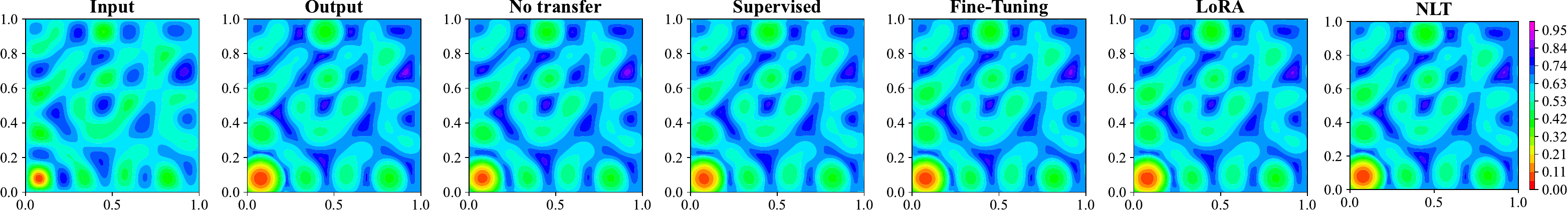}
\end{minipage}

\begin{minipage}{\linewidth}
  \centering
  \includegraphics[width=\textwidth]{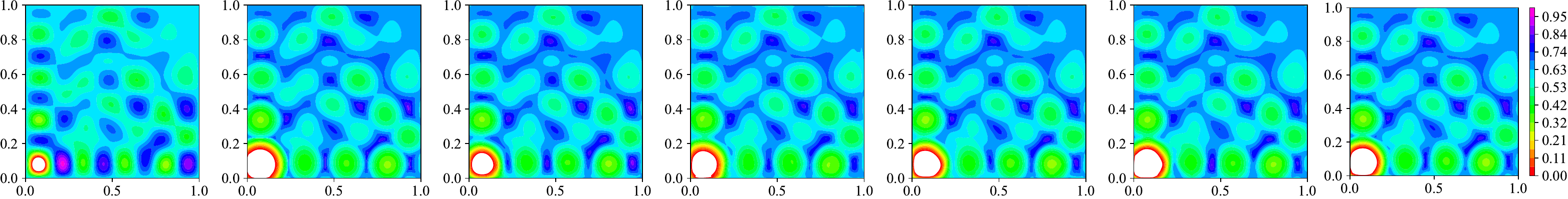}
\end{minipage}

\begin{minipage}{\linewidth}
  \centering
  \includegraphics[width=\textwidth]{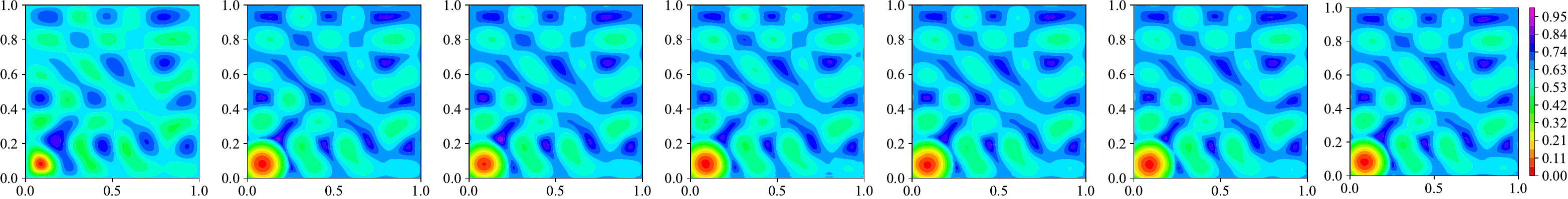}
\end{minipage}

\caption{Contour plots of three test samples for the Kuramoto-Sivashinsky equation for the transfer scenario from $K = 6$ to $K = 8$. The columns display, from left to right: the input function (initial condition $u_0$), the ground truth solution (output function $u(x,T)$), and the predicted solutions from the no-transfer baseline, supervised-learning baseline, fine-tuning, LoRA, and the NLT transfer strategies. Each row corresponds to one of the three test samples in the target dataset. The values of input and output functions are both reshaped into matrices of shape $128\times 128$.}\label{fig2}
\end{figure}

\begin{figure}[H]
\centering

\begin{minipage}{\linewidth}
  \centering
  \includegraphics[width=\textwidth]{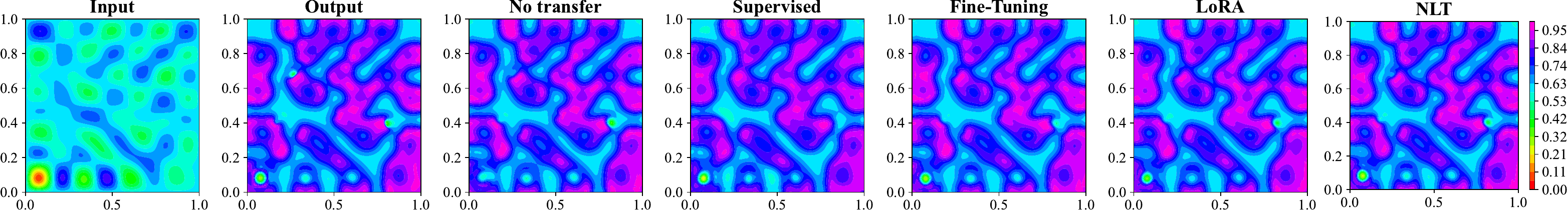}
\end{minipage}

\begin{minipage}{\linewidth}
  \centering
  \includegraphics[width=\textwidth]{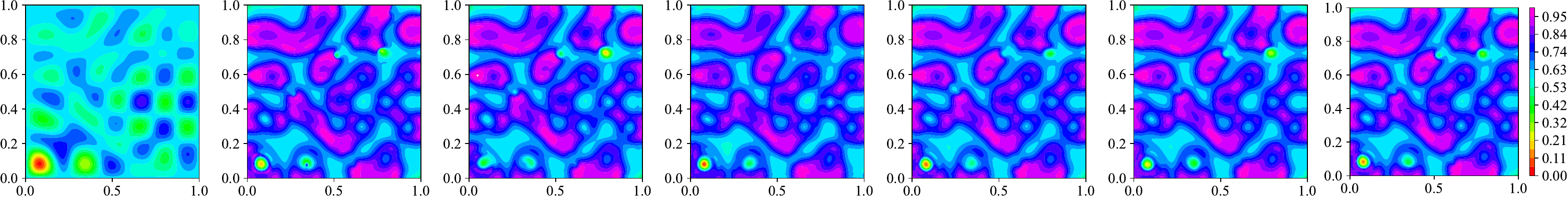}
\end{minipage}

\begin{minipage}{\linewidth}
  \centering
  \includegraphics[width=\textwidth]{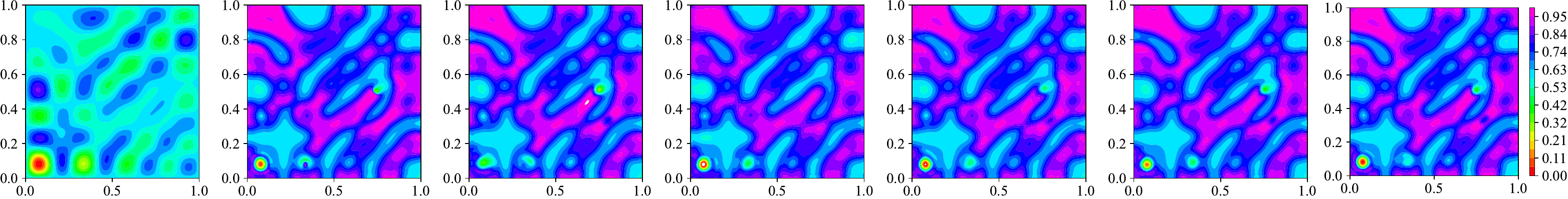}
\end{minipage}

\caption{Contour plots of three test samples for Brusselator diffusion-reaction system for the transfer scenario from $K = 6$ to $K = 8$. The columns display, from left to right: the input function (initial condition $v_0$), the ground truth solution (output function $v(x,T)$), and the predicted solutions from the no-transfer baseline, supervised-learning baseline, fine-tuning, LoRA, and the NLT transfer strategies. Each row corresponds to one of the three test samples in the target dataset. The values of input and output functions are both reshaped into matrices of shape $128\times 128$.}\label{fig3}
\end{figure}

\begin{figure}[H]
\centering
\begin{minipage}{\linewidth}
  \centering
  \includegraphics[width=\textwidth]{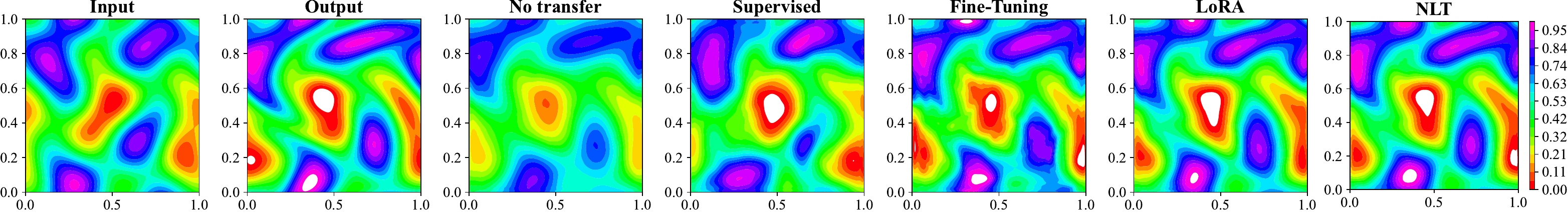}
\end{minipage}

\begin{minipage}{\linewidth}
  \centering
  \includegraphics[width=\textwidth]{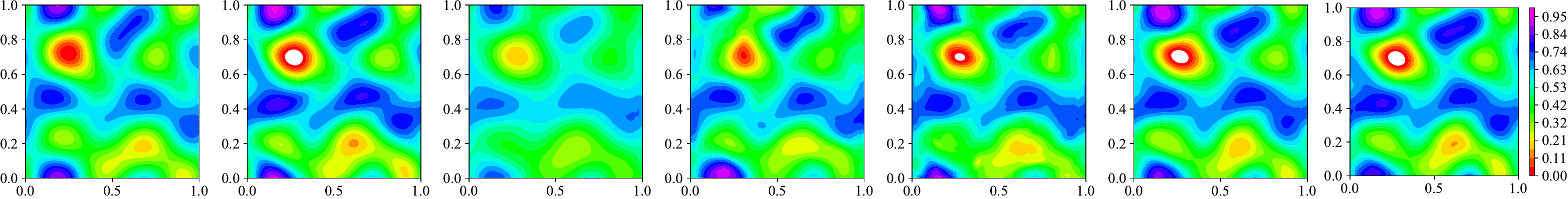}
\end{minipage}

\begin{minipage}{\linewidth}
  \centering
  \includegraphics[width=\textwidth]{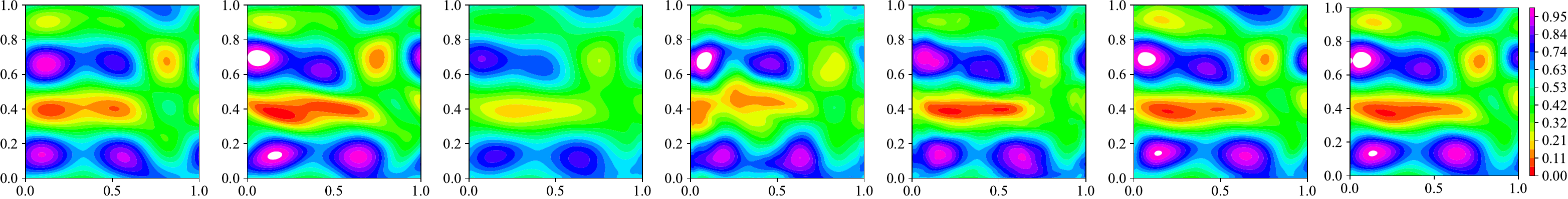}
\end{minipage}
\caption{Contour plots of three test samples for the Navier-Stokes equations for the transfer scenario from viscosity $\nu = 5 \times 10^{-4}$ to $\nu = 1 \times 10^{-4}$. The columns display, from left to right: the input function (initial condition $\omega_0$), the ground truth solution (output function $\omega(x,T)$), and the predicted solutions from the no-transfer baseline, supervised-learning baseline, fine-tuning, LoRA, and the NLT transfer strategies. Each row corresponds to one of the three test samples in the target dataset. The values of input and output functions are both reshaped into matrices of shape $128\times 128$.}\label{fig1}
\end{figure}

 \bibliographystyle{elsarticle-num} 
 \bibliography{ref}

\end{document}